\documentclass[conference]{IEEEtran}

\usepackage{cite}

\ifCLASSINFOpdf
  \usepackage[pdftex]{graphicx}
  \graphicspath{{./images/}}
  \DeclareGraphicsExtensions{.pdf,.jpeg,.png}
\else
  \usepackage[dvips]{graphicx}
  \graphicspath{{./images/}}
  \DeclareGraphicsExtensions{.eps}
\fi

\usepackage{amsmath}

\usepackage{url}

\usepackage{enumitem}
\usepackage{subfig}

\usepackage{hyperref}

\begin{document}

\title{Detecting Noteheads in Handwritten Scores \\ with ConvNets and Bounding Box Regression}

\author{\IEEEauthorblockN{Jan Haji\v{c} jr.}
\IEEEauthorblockA{Institute of Formal and Applied Linguistics
\\Faculty of Mathematics and Physics\\
Charles University\\
Email: hajicj@ufal.mff.cuni.cz}
\and
\IEEEauthorblockN{Pavel Pecina}
\IEEEauthorblockA{Institute of Formal and Applied Linguistics
\\Faculty of Mathematics and Physics\\
Charles University\\
Email: pecina@ufal.mff.cuni.cz}
}

\maketitle

\begin{abstract}
Noteheads are the interface between the written score
and music. Each notehead on the page signifies one note
to be played, and detecting noteheads is thus an unavoidable step
for Optical Music Recognition. Noteheads are 
clearly distinct objects; however,
the variety of music notation handwriting makes noteheads 
harder to identify, and while handwritten music notation symbol {\em classification} 
is a well-studied task, symbol {\em detection}
has usually been limited to heuristics and rule-based systems
instead of machine learning methods better suited to deal with
the uncertainties in handwriting.
We present ongoing work on 
a simple notehead detector using convolutional
neural networks for pixel classification and bounding box regression
that achieves a detection f-score of 0.97 on binary score images
in the MUSCIMA++ dataset, does not require staff removal,
and is applicable to a variety of handwriting styles
and levels of musical complexity.
\end{abstract}

\section{Introduction}
\label{sec:introduction}

Optical Music Recognition (OMR) attempts to extract musical
information from its written representation, the musical score.
Musical information in Western music
means an arrangement of {\em notes} 
in musical time.\footnote{Here, 
the term "note" signifies the {\em musical} object
defined by its pitch, duration, strength, timbre, and onset;
not the written objects: quarter-note, half-note, etc.}
There are many ways in which music notation may encode 
an arrangement of notes, but an elementary rule
is that one note is encoded by one
{\em notehead}.\footnote{An exception would be
"repeat" and "tremolo" signs in orchestral notation.
Trills and ornaments only seem like exceptions if one
thinks in terms of MIDI; from a musician's perspective,
they simply encode some special execution 
of what is conceptually one note.} 

Given the key role noteheads play, 
detecting them -- whether implicitly or explicitly --
is unavoidable for OMR. At the same time, 
if one is concerned only with replayability and not with 
re-printing the input, noteheads
are one of the few music notation symbols that truly need
detecting (i.e., recovering their existence and location) in the score:
most of the remaining musical information can then be framed 
in terms of classifying the noteheads
with respect to their 
properties such as pitch or duration;
bringing one to the simpler territory
of music notation symbol classification.

Music notation defines noteheads so that they are
quickly discernible, and from printed music,
detecting noteheads has been done using segmentation
heuristics such as projections \cite{Fujinaga1988, Bellini2001}
or morphological operators \cite{Fornes2006, Baro2016}.
However, in handwritten music,
noteheads can take on a variety of shapes and sizes,
as illustrated by fig. 1,
and handwriting often breaks the rules of music notation
topology: noteheads may overlap (or separate from) symbols
against the rules. 
Robust notehead detection in handwritten music 
thus invites machine learning.

\begin{figure}
\label{fig:noteheadvariety}
\includegraphics[width=3.5in]{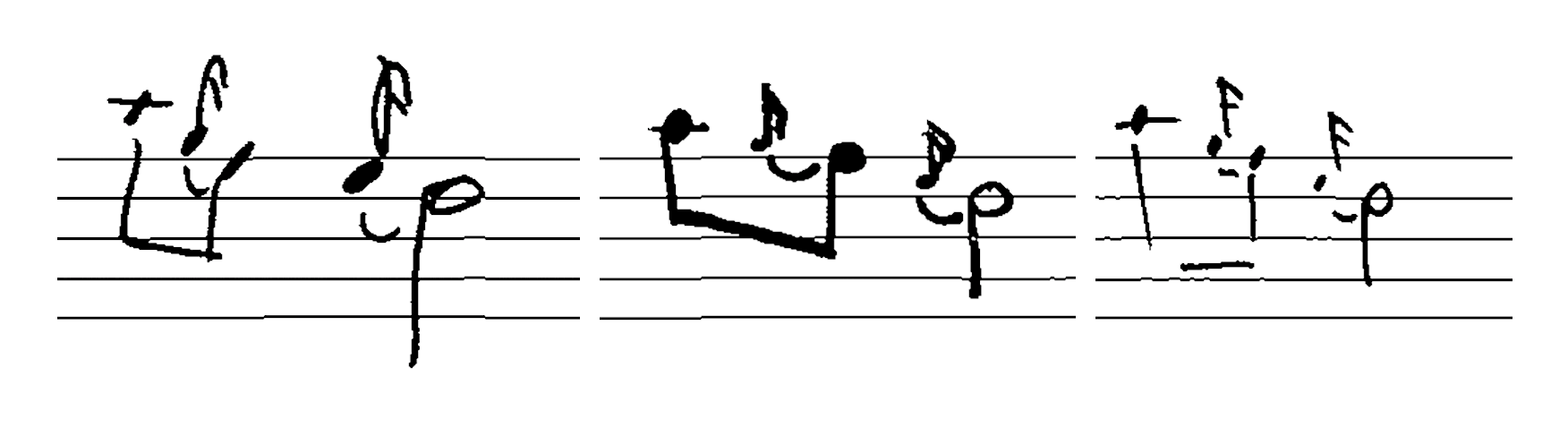}
\caption{The variety of noteheads and handwriting styles: full, empty, and grace
noteheads.}
\end{figure}

{\bf Our contribution is a simple handwritten notehead detector,}
which achieves a detection performance of 0.97 on binary images
across scores of various levels of musical complexity and
handwriting styles. At the heart of the detector is a small
convolutional neural network based on the RCNN family of models,
specifically Faster R-CNN \cite{Ren2015bs}.
Within the traditional OMR pipeline as described by Rebelo et al.
\cite{Rebelo2012}, our work falls within the symbol
recognition stage, specifically as a crucial subset 
of the ``isolating primitive elements''
and jointly ``symbol classification'' steps; 
however, it does {\em not} require staff removal.
In the following sections, we describe the detector in detail,
demonstrate its performance in an experimental setting,
describe its relationship to previous work, and discuss its
limitations and how they can be overcome.

\section{Notehead Detector}
\label{sec:detector}

The notehead detection model consists of three components:
a {\bf target pixel generator} that determines which regions
the detector should attend to,
a {\bf detection network} that jointly
decides whether the target pixel belongs to a notehead 
and predicts the corresponding bounding box, 
and an additional {\bf proposal filter}
that operates on the combined predictions of
the detection network and decides whether
the proposed bounding boxes really correspond to
a notehead.

\subsection{Target Pixel Generator}
\label{subsec:targetpixelgenerator}

The target pixel generator
 takes a binary score image and outputs a set
of $X, t$ pairs, where $X$ is the input for the detection
network corresponding to the location $t = (m, n)$
of a {\em target pixel}. 
From training and validation data,
it also outputs $y = (c, b)$ for 
training the detection network.
The class $c$ is $1$ 
if $t$ lies in a notehead and $0$ otherwise;
$b$ encodes the bounding box of
the corresponding notehead relative to $t$ if $c = 1$
(all values in $b$ are non-negative; they are intepreted as
distance form $t$ to the top edge of the bounding box
of the notehead, to its left edge, etc.);
if $c = 0$, $b$ is set to $(0, 0, 0, 0)$.
The network outputs are described by Fig. 2.

The detection network input $X$ is a patch 
of the image centered on $t$.
The patch must have sufficient size
to capture enough context around a given target pixel,
to give the network a chance to implicitly operate with
rules of music notation, 
e.g. to react to the presence of a stem or a beam in certain positions.
We set the patch size to 101x101 (derived from $1.2 * staff\_height$), 
and downscale to 51x51 for speed.

At {\emph runtime}, we use all pixels of the morphological 
skeleton\footnote{As implemented by the {\tt skimage} Python
library.}
as target pixels $t$.
If one correctly classifies the skeleton pixels and then dilates
these classes back over the score, 
we found over 97 \%
of individual foreground pixels classified correctly
(measured on the MUSCIMA++ dataset \cite{HajicJr2017}),
making the skeleton a near-lossless compression of the score
to about 10 \% of the original foreground pixels.

For {\em training}, we randomly choose $k$ target pixels 
for each musical symbol in the training set,
from the subset of the skeleton pixels
that lies within the given symbol. 
In non-notehead symbols,
we forbid extracting skeleton pixels that are shared 
with overlapping noteheads:
we simply want to know whether a given pixel $t$ belongs
to a notehead or not. 
(This is most pronounced in ledger lines crossing noteheads.)
Setting $k  > 1$ did not improve detection
performance; we suspect this is because all $X$s
from a symbol $S$
are highly correlated and therefore do not give the network 
much new information.

\begin{figure}
\label{fig:extractor}
\includegraphics[width=3.5in]{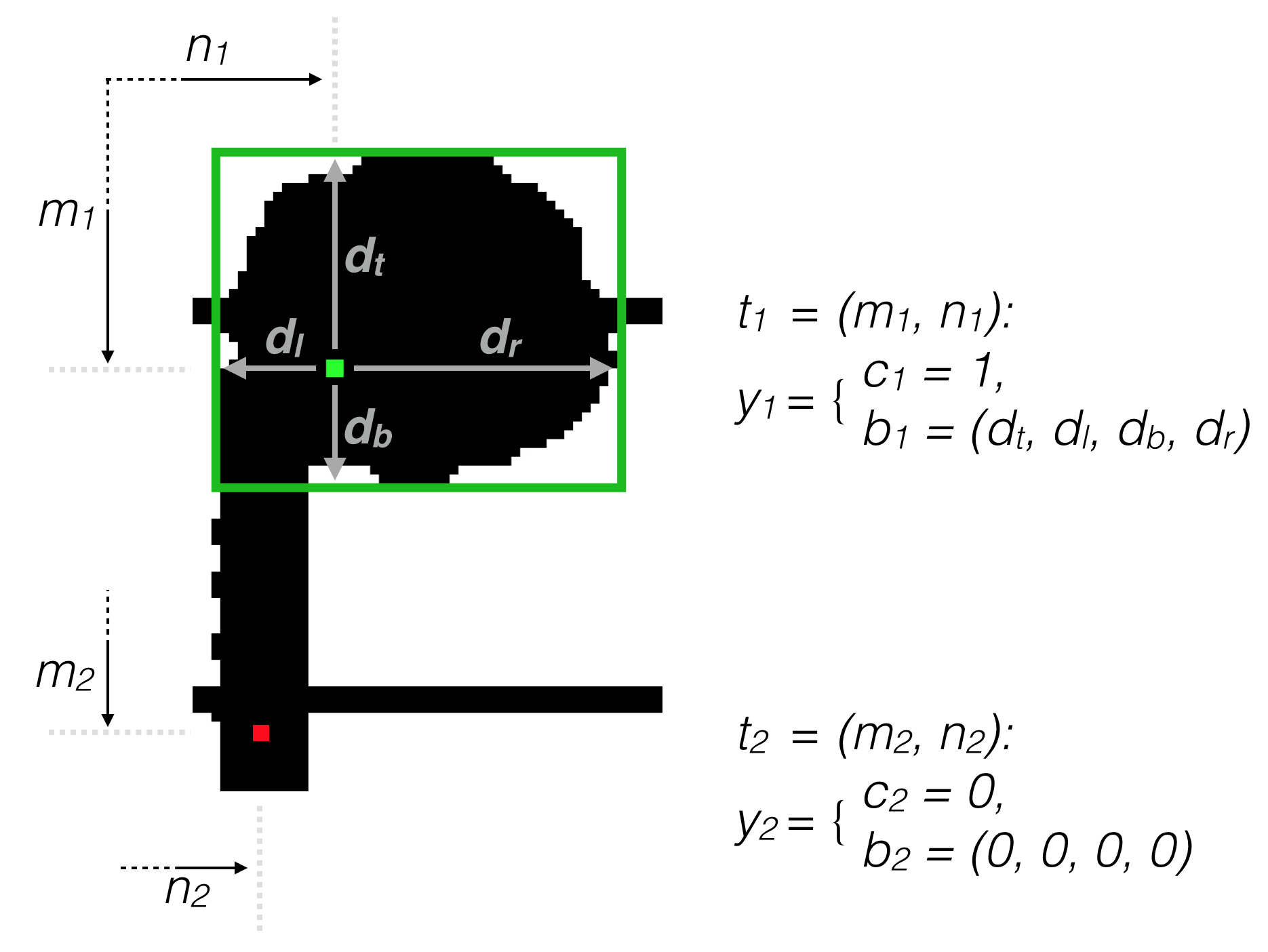}
\caption{The outputs which the detection network is learning for
each target pixel, the green $t_1$ and red $t_2$: 
its class $c$, and the position $b$ of the target pixel
inside the notehead's bounding box -- set to all zeros
when $t$ does not belong to a notehead, as seen in $y_2$.}
\end{figure}

\subsection{Detection Network}
\label{subsec:detectionnetwork}

The detection network handles most of the ``heavy lifting''.
It is a small convolutional network with
two outputs: a binary classifier (notehead-or-not) and
a bounding box regressor. The inputs to the network
are the patches $X_1 \dots X_N$ extracted by
the target pixel generator; the ground truth for training
are the class and bounding box information $y_1 = (c_1, b_1), \dots, y_N$
described in \ref{subsec:targetpixelgenerator}.
This follows the architecture of Faster R-CNN \cite{Ren2015b}, 
but as our inputs are not the natural
images on which VGG16 \cite{Simonyan2014} was trained, we train our own
convolutional stack. (See Sec. \ref{sec:relatedwork} for a more thorough
comparison.)

Our network has four convolutional layers, with the first
two followed by max-pooling with pool size 2x2.
Dropout is set to 0.25 after the max-pooling layers and 0.125 after
the remaining convolutional layers.
The output of the fourth convolutional layer is then
densely connected to the classification output and the bounding
box regression outputs. 
The convolutional layers use $tanh$
activation rather than ReLU: we found that this made learning
converge faster, although we are still unsure why.
Details are given in table \ref{tab:network}.

The classification output uses cross-entropy loss; the bounding
box regression output uses mean squared error loss, weighted
at 0.02 of the classification loss. The network was implemented
using the Keras library \cite{Chollet2015}.

\subsection{Notehead Proposal Filter}
\label{subsec:proposalfilter}

The detection network outputs correspond to individual
target pixels, selected by the generator;
we now combine these results into noteheads.

We take the union
of all bounding boxes output by the detection network
for target pixels with predicted class $c = 1$,
and we use bounding boxes of the connected
components of this union as notehead proposals.
We then train a classifier of notehead proposals. 
This classifier can take into account
all the network's decisions,
as well as other global information;
however, it can only fix false positives, 
 -- if the network misses
a notehead completely, the filter cannot find it. 
However, the detection network in the described setting achieves
good recall and has more trouble with precision,
so such filtering is appropriate.

The features we extract for proposal filtering from each 
proposal region $B_1, \dots, B_j$ are:

\begin{itemize}
\item $h(B)$, the height of $B$,
\item the ratio $\hat{h}(B)$ of $h(B)$ to the average height of $B_1, \dots, B_j$,
\item $w(B)$, the width of $B$, and analogously $\hat{w}(B)$,
\item area $a(B) = h(B) * w(B)$, analogously $\hat{a}(B)$,
\item the no. of foreground pixels in $B$: $N_{fg}(B)$, and the proportion $p_{fg}(B) = N_{fg}(B) / a(B)$,
\item $N^{+}(B)$, the no. of positively classified target pixels $t^{+} \in B$,
\item $p^{+}(B)$, the proportion of such target pixels to all in $t^{+} \in B$,
\item the ratio $\hat{N}^{+}(B)$,
\item equiv. for non-notehead pixels $t^{-} \in B$: $N^{-}(B)$, $p^{-}(B)$, $\hat{N}^{-}(B)$, 
\item "soft" sum of noteheadedness: $S^{+}(B) = \sum_{t^{+} \in B} P(+ \mid t)$,
\item again, the ratio to the average $S^{+}(B)$ in the image: $\hat{S}^{+}(B)$.
\item $l(B)$: how much to the left in the input image $B$ is.
\end{itemize}

\noindent
The ratio features ($\hat{w}(B)$, etc.) are designed to simulate
invariance to individual handwriting styles.
Also, beyond features based on detection network outputs,
the left-ness $l(B)$ is used to find false positives in clefs.

For training the proposal filter, 
we consider correct each notehead proposal 
that has Intersection-over-Union (IoU)
with a true notehead above 0.5.
We use a Random Forest with 300 estimators, a maximum
depth of 8, and a minimum of 3 samples in each leaf.

\section{Experiments}
\label{sec:experiments}

We now describe the experimental setup in which
the detector was tested: 
the dataset, evaluation procedure,
and experimental results.

\subsection{Data}
\label{subsec:experiments:data}

For experiments, we use the MUSCIMA++ dataset of
Haji\v{c} jr. and Pecina \cite{HajicJr2017}, based on the underlying
images of CVC-MUSCIMA by Forn\'{e}s et al. \cite{Fornes2012}. 
The dataset contains 140 binary images. There are 20 pages of music,
each as transcribed by 7 of the 50 writers of CVC-MUSCIMA;
all the 50 CVC-MUSCIMA writers are represented in MUSCIMA++.
The scores all use the same staffline and staffspace heights
(see \cite{Fornes2012} for details).
We use a test set that contains one of each
of the 20 pages, chosen so that no page by the writers of the test
set pages is seen in the training set (we first want to see how
the system generalizes to unseen handwriting style, rather than
unseen notational situations).
When extracting ground truth, we did not differentiate between different 
noteheads (full, empty, grace-note).

We used the first 100 of the remaining 120 images as the training
set for the detection network, and the other 20 as the validation set.
As we are training on only one sample target pixel per musical
symbol, this amounted to 65015 training instances.
Using the Adam optimizer,
training converged after 8-9 epochs.

The outputs of the network on the dev set 
were then included for training the notehead proposal filter, together
with the first 50 images from the training set. 
(The dev set better approximates the inputs
to the proposal filter at runtime conditions, when the detection
network runs on images never seen in training.) 

\begin{table}
\centering
  \label{tab:network}
  \begin{tabular}{lccl}
    \hline
    Layer & Dropout & Activation & Size \\
    \hline
    conv1 & -- &  tanh & 32 filters 5x5 \\
    pool1 & 0.25 & -- & pool 2x2 \\
    conv2 & -- & tanh &   64 filters 3x3 \\
    pool2 & 0.25 & -- & pool 2x2 \\
    conv3 & 0.125 & tanh &   64 filters 3x3 \\
    conv4 & 0.125 & tanh &   64 filters 3x3 \\
    \hline
    clf. & -- & sigmoid & 1 \\
    bb. reg. & -- & ReLU & 4 \\
  \hline
\end{tabular}
  \caption{Detection Network Architecture. 
 (Both the classification and bounding box regression output layers are densely connected to conv4.)}
\end{table}

\subsection{Evaluation Procedure}
\label{subsec:experiments:evaluation}

We evaluate notehead detection recall and precision.
A notehead prediction
that has IoU over 0.5 with a true
notehead is a hit. Furthermore, we count each predicted notehead 
that completely contains a true notehead.
This non-standard way of counting hits was chosen
because in some cases, the bounding box regression produced bounding boxes
that were symmetrically "around" the true notehead,
but slightly too large, to the extent that it set
IoU too low due to the predicted notehead's contribution to the union term. 
However, a symmetrically larger bounding box 
(when oversized only to the limited
extent present in the model outputs) 
does not impede recovering the notehead's relationship
to other musical symbols, e.g., stafflines, and
this adjustment should therefore give the reader 
a better grasp of the detector's actual useful
performance.\footnote{Technically, this adjustment moved recall 
upwards by 3 - 5 \% across all images.}

\subsection{Results}
\label{subsec:experiments:results}

{\bf On average, the detector achieves a recall of 0.96 and precision of 0.97.}
Among the test set,
there were two images where recall fell to around 0.9:
0.87 for CVC-MUSCIMA image W-12\_N-19 
(writer 12, page 19), and 0.91 for W-29\_N-10, 
due to the detection network's 
errors on empty noteheads 
in the middle of chords, full noteheads
in chords with a handwriting style 
where the notehead is essentially just 
a thickening or straight extension of the stem, 
and certain grace notes; 
there are also problems with whole notes
on the ``wrong'' side of the stem in W-39\_N-20. 
Aside from these situations, 
the detector rarely misses a note.\footnote{Visualizations 
of results for the test set images are available online: 
\url{https://drive.google.com/open?id=0B9l5xUyYe-f8Y2FQWTZxc09PaEE}}

The detection network itself has an average {\em pixel-wise}
recall on the positive class of 0.94 (again, the average is lowered
mostly by the three problematic images, rather than evenly distributed
errors), but precision only 0.78 (even though most of the false positives
are skeleton pixels in the close vicinity of actual noteheads). 
The {\em notehead detection} 
recall without post-filtering is 0.97 and precision is 0.81.
As the false positives are clearly a much greater problem than false negatives,
preliminary results to this effect on the development set 
motivated work on the post-filtering step.
The post-filter increases
detection precision by 0.16, eliminating over 84 \% of all the false positives,
while only introducing 1 \% more false negatives.

\section{Related work}
\label{sec:relatedwork}

Given that there was little publicly available ground truth
for notehead detection until recently \cite{HajicJr2017},
it is hard to compare results to previous work directly.
A noteworthy approach on the same CVC-MUSCIMA
handwritten data was taken by Baro et al. \cite{Baro2016}. 
They achieve a notehead detection f-score of 0.64 based on
handcrafted rules alone, without any machine learning.
This is an indication
that contemporary handwritten music 
will need a machine-learning approach
rather than the projection-based heuristics that have been used 
in printed music \cite{Fujinaga1988, Bellini2001} 
and applied to handwritten early music scores 
with recall 0.99 and precision 0.75 \cite{Fornes2006}.

Convolutional neural networks have been previously successfully
applied to music scores by Calvo-Zaragoza et al.,
for segmentation into staffline, notation,
and text regions 
\cite{CalvoZaragoza2017b} or binarization
\cite{CalvoZaragoza2017}, with convincing results that generalize
over various input modes.

Our detector is inspired by 
the RCNN family of models, especially Faster R-CNN
\cite{Ren2015b}. 
RCNNs were
motivated by the fact that detection can be decomposed
into region proposals and classification, with
models such as VGG16 \cite{Simonyan2014} for natural
image classification obtaining near-human performance. 
However, the pre-trained image classification nets are
too slow for a trivial sliding window approach. 
RCNNs use a sparse grid
of proposal regions with pre-defined sizes and shapes, 
and train bounding box regression to locate
the object of interest within the proposal region.
(Faster R-CNN trains bounding box regression directly
on top of the high-quality
image classification features.)
When combining predictions to obtain detection outputs,
RCNN models then apply non-maximum suppression
on the detection probability landscape obtained from
predictions for each of the pre-defined
proposal regions.

Together with \cite{Ren2015b}, we apply joint classification and bounding box
regression, but our approach differs from RCNNs in four aspects.
We do not use a fixed proposal grid but generate
proposal regions dynamically from the input image.
Second, we cannot reuse VGG16 \cite{Simonyan2014} or other powerful
pre-trained image classification models for feature extraction, 
since they not trained on music notation data; 
however, because our input space is much simpler,
we can train the convolutional layer stack directly.
Third, we use a separate classifier to take advantage
of the network outputs for related proposal regions,
combining the network's "votes" on multiple closely related inputs
more generally than simply non-maxima suppression.
A final subtle distinction is that we are not just looking
for a notehead anywhere in the proposal region;
we want the {\em center pixel} of the region to be part
of the notehead, constraining bounding box regression
outputs roughly to the average size of a notehead even with
a much larger input patch.

\section{Discussion and Conclusions}
\label{sec:conclusions}

We proposed an accurate notehead detector from simple 
image processing and machine learning components. 
However, ongoing work on the detector will need
to address several limitations.

Our system requires binary images.
The detection network can be trained on augmented grayscale data,
and given the track record of convolutional networks, one would expect
good performance; however, an alternate target pixel selection mechanism
is needed.

A second problem is slow runtime: 
over a minute per MUSCIMA++ test set image
on a consumer
CPU. This can be mitigated by first downsampling the skeleton, and
then informing the choice of more target pixels by the results,
directing the network to focus only on "hopeful" areas where
it has detected a notehead.

While the binary nature of the 
task is appealing, the network is in fact forced to lump
different symbols together. This is more pronounced
in the negative class, where the variety of shapes is larger.
Saliency maps for the last convolutional layer
suggest that most of its filters relate to the presence of a stem;
forcing the network to discriminate among more classes might force
convolutional filters to distinguish specifically between noteheads
and similar objects.

The final issue is generalizing 
past the high-quality scans of CVC-MUSCIMA images. 
In preliminary experiments on an early music page 
with low-quality binarization and some
blurring and deformation,
the detector gets f-score 0.85 without and 0.79 with
post-filtering. The more fragile stages of the detector are at fault --
a low-quality skeleton, and the correspondingly uncertain inputs
for the post-filtering classifier. (The post-filtering classifier is fragile
in the sense that its features are directly derived from the combined outputs
of the detection network, and thus it is ``conditioned'' for a certain
level of network performance.)

These limitations suggest a way forward:
more efficient target pixel selection, applicable to grayscale images;
data augmentation to simulate more real-world conditions;
a more robust post-filtering step, ideally trainable
jointly with the detection network;
and extending the detection network to multiple output classes
(which, when combined with using previous outputs 
in the vicinity of a given target pixel as an input to the network,
can also incorporate music notation syntax more explicitly).

The simple detector has proven to be quite powerful, 
resistant to changes in handwriting style and most notation complexity, 
showcasing the potential of quite simple
neural networks (and the value of a dataset). 
In spite of the limitations,
we find this an encouraging result for offline handwritten OMR.

\section*{Acknowledgments}

This work is supported by the Czech Science Foundation,
grant number P103/12/G084, the Charles University Grant Agency,
grants number 1444217 and 170217, and by SVV project number 260 453.



\bibliographystyle{IEEEtran}
\bibliography{bibliography}
%
%
%

\end{document}